\documentclass{ecai} 

\usepackage{latexsym}
\usepackage{amssymb}
\usepackage{amsmath}
\usepackage{amsthm}
\usepackage{booktabs}
\usepackage{enumitem}
\usepackage{graphicx}
\usepackage{color}

\usepackage{pgf}
\usepackage{svg}
\usepackage{soul}
\setstcolor{red}

\usepackage{todonotes}
\usepackage{xspace}
\usepackage{amsmath,amsthm,amssymb}
\usepackage{thmtools,mathtools}

\usepackage{anyfontsize} 

\usepackage{multirow}
\usepackage{booktabs}

\usepackage{algorithm}
\usepackage{algorithmicx}
\usepackage[noend]{algpseudocode}

\algrenewcommand\alglinenumber[1]{\scriptsize #1:}
\MakeRobust{\Call}

\newcommand{\algsetstretch}{\linespread{.9}\selectfont}

\usepackage[noabbrev, capitalize]{cleveref}

\usepackage{tikz}
\usetikzlibrary{positioning, shapes, calc, chains, decorations.pathreplacing, backgrounds}

\usepackage{relsize}
\tikzset{fontscale/.style = {font=\relsize{#1}}}

\usepackage{enumitem}
\setitemize{noitemsep,topsep=0pt,parsep=0pt,partopsep=0pt,leftmargin=0pt,itemindent=10pt}
\setenumerate{noitemsep,topsep=0pt,parsep=0pt,partopsep=0pt,leftmargin=0pt,itemindent=10pt}
\usepackage{scalefnt}

\newcommand{\dataset}{\Tilde{\mathcal{P}}^k_{\mathcal{D}}}

\newcommand{\dummyaction}{\ensuremath{\bar{m}}\xspace}
\newcommand{\dummystate}{\ensuremath{\bar{d}}\xspace}
\newcommand{\deadends}{S_d}
\newcommand{\goalstates}{S_g}
\newcommand{\appl}[1]{\alpha(#1)}
\newcommand{\successor}[2]{\nu(#1,#2)}  

\newcommand{\tuple}[1]{\langle #1 \rangle}
\newcommand{\eqbydef}{\: \dot{=} \:}

\newcommand{\initdistrib}{\ensuremath{I}\xspace}

\newcommand{\deltarl}{\ensuremath{\Delta_{RL}}\xspace}
\newcommand{\deltahold}{\ensuremath{\Delta_H}\xspace}
\newcommand{\deltahnew}{\ensuremath{\Delta_h}\xspace}

\newcommand{\oldr}{\ensuremath{R_{bin}}}
\newcommand{\newr}{\ensuremath{R_{cnt}}}
\newcommand{\oldv}{V_{bin}}
\newcommand{\newv}{V_{cnt}}
\newcommand{\oldh}{\ensuremath{h_{bin}}}
\newcommand{\newh}{\ensuremath{h_{cnt}}}
\newcommand{\oldres}{\ensuremath{r_{bin}}}
\newcommand{\newres}{\ensuremath{r_{cnt}}}
\newcommand{\oldphi}{\ensuremath{\phi_{bin}}}
\newcommand{\newphi}{\ensuremath{\phi_{cnt}}}

\newcommand{\newphip}{\ensuremath{\phi_{cnt}'}}
\newcommand{\rank}{\ensuremath{u}}

\newcommand{\mymdp}{\mathcal{M}[\mathcal{P}_\mathcal{D}^k]\xspace}
\newcommand{\myoldmdp}{\mathcal{M}_{bin}\xspace}
\newcommand{\mynewmdp}{\mathcal{M}_{cnt}\xspace}
\newcommand{\myhstar}{h^*_{\mathcal{P}_\mathcal{D}^k}}

\newcommand{\symh}{\ensuremath{h_{sym}}\xspace}

\newcommand{\sigmoid}{\omega}

\newtheorem{definition}{Definition}
\newtheorem{proposition}{Proposition}
\newcommand{\deftitle}[1]{\textbf{#1}\xspace}

\newcommand{\sstodo}{\lambda\xspace}
\newcommand{\sstn}{\chi\xspace}
\newcommand{\sslast}{\omega\xspace}

\newcommand{\majsp}{\ensuremath{\textsc{MAJSP}}\xspace}

\newcommand{\tamer}{\textsc{Tamer}\xspace}

\usepackage{listings}

\definecolor{dkgreen}{rgb}{0,0.5,0}
\definecolor{dkblue}{rgb}{0,0,0.5}
\lstdefinelanguage{ANML}
{
  keywords={},
  keywords=[2]{
    action, coincident, comprises, constant, contains,  else, exists,
    fact, fluent, forall, function, goal, instance, motivated, ordered,
    predicate, type, unordered, variable, when, with, duration
  },
  keywords=[3]{
    all, start, end
  },
  keywords=[4]{
    boolean, integer, string, float, set
  },
  keywords=[5]{
    true, false, UNDEFINED, infinity
  },
  keywords=[6]{
    in, xor, not, or, implies, and, iff, union, intersect, subset, powerset, elt
  },
  otherkeywords={
    :produce, :in, :use, :decomposition, :lend, :=, :consume, :->, :subset
  },
  morecomment=[l]{//},
  morecomment=[s]{/*}{*/},
  morestring=[b]"
  sensitive=true,
  breaklines=true,
  extendedchars=true,
  basicstyle=\scriptsize\ttfamily,
  numberstyle=\tiny,
  numbers=left,
  numbersep=.3cm,
  columns=fixed,
  xleftmargin=.6cm,
  framexleftmargin=.5cm,
  showstringspaces=false,
  showtabs=false,
  commentstyle=\color{dkgreen},
  keywordstyle={[1]\color{dkblue}},
  keywordstyle={[2]\color{dkblue}\bf\ttfamily},
  keywordstyle={[3]\color{magenta}},
  keywordstyle={[4]\color{dkblue}},
  keywordstyle={[5]\color{red}},
  keywordstyle={[6]\color{red}}
}

\lstnewenvironment{anml}{\lstset{language=ANML, aboveskip=0.2cm, belowskip=0.2cm}}{}

\begin{document}


\begin{frontmatter}


\paperid{4770} 


\title{Exploiting Symbolic Heuristics for the Synthesis of Domain-Specific Temporal Planning Guidance using Reinforcement Learning}


\author[A]{\fnms{Irene}~\snm{Brugnara}\thanks{Corresponding Author. Email: ibrugnara@fbk.eu.}}
\author[A]{\fnms{Alessandro}~\snm{Valentini}}
\author[A]{\fnms{Andrea}~\snm{Micheli}} 

\address[A]{Fondazione Bruno Kessler, Trento, Italy}


\begin{abstract}
Recent work investigated the use of Reinforcement Learning (RL) for the synthesis of heuristic guidance to improve the performance of temporal planners when a domain is fixed and a set of training problems (not plans) is given. The idea is to extract a heuristic from the value function of a particular (possibly infinite-state) MDP constructed over the training problems. 

In this paper, we propose an evolution of this learning and planning framework that focuses on exploiting the information provided by symbolic heuristics during both the RL and planning phases. First, we formalize different reward schemata for the synthesis and use symbolic heuristics to mitigate the problems caused by the truncation of episodes needed to deal with the potentially infinite MDP. Second, we propose learning a residual of an existing symbolic heuristic, which is a ``correction'' of the heuristic value, instead of eagerly learning the whole heuristic from scratch. Finally, we use the learned heuristic in combination with a symbolic heuristic using a multiple-queue planning approach to balance systematic search with imperfect learned information.
We experimentally compare all the approaches, highlighting their strengths and weaknesses and significantly advancing the state of the art for this planning and learning schema.
\end{abstract}


\end{frontmatter}


\section{Introduction}

Automated temporal planning is the problem of synthesizing a course of action to achieve a desired goal condition, given a formal description of a timed system~\cite{traverso-book}. Temporal planning finds natural applications in many domains such as robotics, logistic and process automation because time and temporal constraints are commonplace in real-world applications. Despite a long history of research and improvements, scalability is the key issue for automated temporal planners: depending on the modeling assumptions the computational complexity ranges from PSPACE-completeness to undecidability~\cite{gigante-aij}. Using machine learning to specialize a temporal planner for a domain is a promising way to mitigate the scalability issue and automatically devise effective solvers.

An approach to tackle this specialization of temporal planning solvers consists of codifying a set of training problems as a specifically designed Markov Decision Process (MDP) and using Reinforcement Learning~\cite{sutton-barto} to estimate its value function; in turn, it is possible to extract a temporal planning heuristic biased on the distribution of training problems for a search-based temporal planner from the estimated value function~\cite{aaai21}.
The workflow consists of two phases: an offline phase in which the temporal planner is specialized on the given training dataset in a fully automated way, and then an online phase in which the specialized planner is tasked to solve multiple problems on the same domain.
More in detail, this approach builds on the discretization of the search space used by planners such as \tamer \cite{tamer}: the continuous space of the timed system is abstracted by symbolically representing time and timing constraints using Simple Temporal Networks (STN) \cite{stn} and the planner searches in the (partial) orderings of \emph{events}, which are either the beginning or ending of durative actions, or intermediate effects and conditions if they are allowed. The learning framework exploits this discretization of the search space by defining a tree-shaped MDP%
, where an instance from the training set is randomly selected at the beginning of the episode and then the MDP is isomorphic to the planner search space. An optimal value function for the MDP is shown to be in a functional relation with the optimal heuristic for the training instances, and the authors show that by using neural networks as function approximators and encoding states (including symbolic temporal information and the goal) as feature vectors, the approach can generalize to instances drawn from the same distribution of the training set.

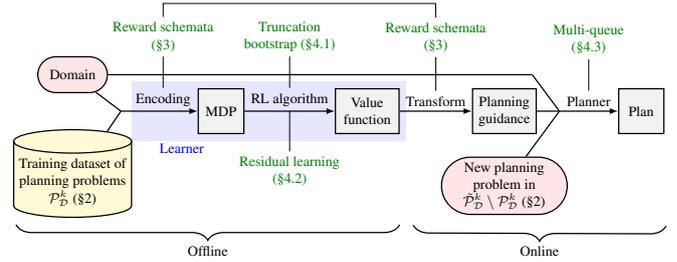
\begin{figure}[tb]
    \centering
    \resizebox{\columnwidth}{!}{\begin{tikzpicture}
  [
    node distance=.8cm,
    start/.style = {circle, draw=black, fill=red!10, thick, minimum size=8mm},
    instance/.style = {rounded rectangle, draw=black, fill=red!10, thick, minimum size=8mm},
    artifact/.style = {rectangle, draw=black, fill=gray!10, thick, minimum size=10mm},
    tool/.style = {rectangle, draw=black, fill=green!10, thick, minimum size=15mm},
    service/.style = {rectangle, draw=black, thick, minimum width=20mm},
    callout/.style = {text={rgb, 255:red,0; green,128; blue,0}},
    link/.style   ={thick, ->, >=latex},
    line/.style   ={thick, -},
    dataset/.style = {
    cylinder,
    shape border rotate=90,
    draw=black,
    thick,
    shape aspect=0.2,
    minimum height=16mm,
    minimum width=15mm,
    cylinder uses custom fill,
    cylinder body fill=yellow!20,
    cylinder end fill=yellow!20
    }
  ]
    \pgfdeclarelayer{background}
    \pgfsetlayers{background,main}

    \node[instance] (domain) {\parbox{1.3cm}{\centering Domain}};
    \node[dataset] (training)  [below=of domain] {\parbox{2.4cm}{\centering Training dataset of planning problems $\mathcal{P}_\mathcal{D}^k$ (\S \ref{section:problem})}};
    \node[inner sep=0pt, outer sep=0pt, minimum size=0pt] (encoding-start) [below right=0.4cm and 0.7cm of domain] {};
    \node[artifact] (MDP)  [right=1.7 cm of encoding-start] {\parbox{0.8cm}{\centering MDP}};
    \node[artifact] (V) [right= 2cm of MDP] {\parbox{1.2cm}{\centering Value function}}; 
    \node[artifact] (guidance) [right=1.6cm of V] {\parbox{1.2cm}{\centering Planning guidance}};  
    \node[instance] (problem) [below=0.5cm of guidance] {\parbox{2cm}{\centering New planning problem in $\dataset \setminus \mathcal{P}_\mathcal{D}^k$ (\S \ref{section:problem})}};
    \node[inner sep=0pt, outer sep=0pt, minimum size=0pt] (planner-start) [right=0.5cm of guidance] {};
    \node[inner sep=0pt, outer sep=0pt, minimum size=0pt] (domain-end) [right=9.4cm of domain] {};
    \node[artifact] (plan) [right=1.8cm of guidance] {\parbox{0.6cm}{\centering Plan}};

    \begin{pgfonlayer}{background}
        \fill[color=blue!10] ($ (MDP.north west) + (-1.45, 0.15) $) node[above,anchor=center, xshift=1.1cm, yshift=-1.5cm,color=blue]{ Learner} rectangle ($ (V.south east) + (0.15, -0.15) $);
    \end{pgfonlayer}

    \draw (domain) edge[line] (encoding-start);
    \draw (training) edge[line] (encoding-start);
    \draw (encoding-start) edge[link] node [pos=0.55, above] (encoding) {Encoding} (MDP);
    \draw (MDP) edge[link] node [midway, above] (algo) {RL algorithm} (V);
    \draw (V) edge[link] node [midway, above] (transform) {Transform} (guidance);
    \draw (domain) edge[line] (domain-end);
    \draw (guidance) edge[line] (planner-start);
    \draw (domain-end) edge[line] (planner-start);
    \draw (problem) edge[line] (planner-start);
    \draw (planner-start) edge[link] node [midway, above] (planner) {Planner} (plan);
    

    \node[callout] (callout-encoding) [above=0.7cm of encoding] {\parbox{2.3cm}{\centering Reward schemata (\S \ref{section:mdp})}};
    \draw (callout-encoding) edge (encoding);

    \node[callout] (callout-algo1) [below=0.9 of algo] {\parbox{2.3cm}{\centering Residual learning (\S \ref{section:residual})}};
    \draw (callout-algo1) edge (algo);

    \node[callout] (callout-algo2) [above=0.7cm of algo] {\parbox{2.3cm}{\centering Truncation bootstrap (\S \ref{section:bootstrap})}};
    \draw (callout-algo2) edge (algo);
    
    \node[callout] (callout-transform) [above=0.75cm of transform] {\parbox{2.3cm}{\centering Reward schemata (\S \ref{section:mdp})}};
    \draw (callout-transform) edge (transform);


    \node[callout] (callout-planner) [above=0.7cm of planner] {\parbox{2.3cm}{\centering Multi-queue (\S \ref{section:multiqueue})}};
    \draw (callout-planner) edge (planner);

    \draw[]
    (callout-encoding.north) -- ++(0,0.3) -- ++(6.04,0) -- (callout-transform.north);

     \draw[decorate, decoration={brace,amplitude=10pt}, thick] ($ (training.south west) + (8, -0.2) $) -- node[midway, below=10pt]{Offline} ($ (training.south west) + (-0.5, -0.2) $);
     \draw[decorate, decoration={brace,amplitude=10pt}, thick] ($ (V.south east) + (5.9, -2.05) $) -- node[midway, below=10pt]{Online} ($ (V.south east) + (0.3, -2.03) $);

\end{tikzpicture}}
    
    \caption{Overall learning and planning framework. In green we indicate our major contributions.}%
    \label{fig:overall-schema}%
\end{figure}

In this paper, we evolve this framework by leveraging information from existing symbolic temporal planning heuristics as shown in \Cref{fig:overall-schema}.  
We start by considering different reward schemata for the framework (all capturing the temporal plan validity) and by defining a ``truncation bootstrap'' method to cope with the problem of truncation of episodes: since the MDP derived from a temporal planning problem is potentially infinite, we need to truncate episodes after a certain number of steps, and we use the symbolic heuristic to provide a value function estimation in the last state of the episode avoiding learning issues and allowing for better generalization.
Then, we propose to reframe the learning problem from the eager synthesis of a heuristic function to the learning of an additive residual over an existing symbolic heuristic: the key intuition is to simplify the learning task by providing a template function to correct instead of eagerly regressing the whole function from scratch, resulting in a more regular target function.
Finally, we modify the inference phase employing a multiple-queue planning approach, where one queue is ordered using a standard symbolic heuristic in an $A^*$ fashion, while the other is ordered using the learned value function as a ranking function. The intuition is that the second queue exploits the learned information greedily, while the first queue provides a systematic exploration of the search space and compensates for learning imperfections easily compared to a single-queue $A^*$ approach where the systematic search is controlled by the cumulative cost ($g$).
The overall result is a comprehensive learning and planning framework for temporal planning that leverages existing symbolic heuristics at every step.    

We empirically evaluate all components of the modified framework in isolation and in combination, showing they dramatically outperform the baseline planner with no learned information and significantly improve the results in \cite{aaai21}. 
In this setting of per-domain learning, we are interested in optimizing the online solving time, and not the offline learning time (provided it remains feasible).


\section{Problem Definition} \label{section:problem}

In this section, we define the learning problem we tackle in this paper. We borrow most of the basics of our problem definition from \cite{aaai21}.
We use ANML~\cite{anml} to specify action-based temporal planning problems: our language allows durative actions (similarly to PDDL 2.1 \cite{pddl21}) and "Intermediate Conditions and Effects", allowing actions to check for conditions and apply effects at arbitrary points in time during action execution. For the sake of brevity, we do not report the details of the formal planning language here; the details can be found in \cite{tamer}. 
For this paper, we only need to characterize the set of planning problems we use for training and the search space of a planner on such problems.
As customary in planning, we define a planning instance as the composition of a ``domain'' specification (containing the set of predicates and action schemata) and a ``problem'' consisting of a set of objects to instantiate the domain elements, an initial state and a goal condition.

\begin{definition}
  A \deftitle{planning problem} $\mathcal{P}$ for a domain $\mathcal{D}$ is a tuple $\tuple{\mathcal{O}, \mathcal{I}, \mathcal{G}}$ where $\mathcal{O}$ is a finite set of objects; 
  $\mathcal{I}$ and $\mathcal{G}$ are sets of ground atoms over predicates of $\mathcal{D}$, which represent the initial state and the goal respectively.
  A \deftitle{planning instance} is a pair $\tuple{\mathcal{D}, 
  \mathcal{P}}$.
\end{definition}

Semantically, given a planning instance we can first perform standard grounding, that is get rid of action parameters by substituting all the possible combinations of objects as parameters. Given the ground planning instance, a valid plan is a simulation of the system starting from the initial state and terminating in a goal state. 
The full semantics can be found in \cite{tamer}.

Given a ground planning instance, planners such as POPF \cite{popf} or \tamer \cite{tamer} search an interleaving of \emph{events} (also called happenings, time-points or snap-actions) that represent the discrete changes of state in a plan ensuring that the abstract sequence of events can be concretized to a plan by scheduling the temporal constraints.
The timings of events are derived by solving a constraint satisfaction problem (a Simple Temporal Network \cite{stn}, in particular) composed of the temporal constraints and the precedence constraints enforcing causal relations that the planner builds while constructing the sequence of events.
In order to construct such a sequence of events and to record the temporal constraints, we represent search states as per definition below, borrowed from \cite{tamer}, and we perform a search in the space of the possible reachable states starting from the initial state.
The transitions considered by the planner are the events for a planning problem $\mathcal{P}$ (indicated as $\mathcal{E}_{\tuple{\mathcal{D}, \mathcal{P}}}$) and are either instantiations of new actions or expansions of time-points, each indicating an effect, the starting of a condition or its ending.
\begin{definition}
  \label{def:search-state}
  A \textbf{search state} is a tuple $\tuple{\mu, \delta, \sstodo, \sstn, \sslast}$ s.t.:
  \begin{itemize}
  \item $\mu$ records the ground predicates that are true in the state;
  \item $\delta$ is a multiset of ground predicates, representing the active durative conditions to be maintained valid;
  \item $\sstodo$ is a list of lists of time-points. It constitutes
    the ``agenda'' of future commitments to be resolved.
  \item $\sstn$ is a Simple Temporal Network (STN) defined over time-points that stores and checks the metric and precedence temporal constraints;
  \item $\sslast$ is the last time-point evaluated in this search branch.
  \end{itemize}
\end{definition}

We indicate the (infinite) set of possible search states for a given instance $\tuple{\mathcal{D}, \mathcal{P}}$ as $\mathcal{S}_{\tuple{\mathcal{D}, \mathcal{P}}}$.
For deterministic temporal planning, the graph whose set of nodes is $\mathcal{S}_{\tuple{\mathcal{D}, \mathcal{P}}}$ and whose set of edges is $\mathcal{E}_{\tuple{\mathcal{D}, \mathcal{P}}}$ is a tree rooted at the state corresponding to $\mathcal{I}$. This is because most planners do not perform state merging\footnote{If state merging is performed in temporal planning, our algorithms would also work on the DAG-shaped search space; but not in the classical planning case, where the search space is a generic graph.} as it is computationally heavy for temporal planning in general \cite{temporal-subsumption}.
Given a state $\sigma \in \mathcal{S}_{\tuple{\mathcal{D}, \mathcal{P}}}$, we denote with $\appl{\sigma}$ the set of applicable events in $\sigma$. Given a state $\sigma \in \mathcal{S}_{\tuple{\mathcal{D}, \mathcal{P}}}$ and an event $\epsilon \in \appl{\sigma}$, we denote with $\successor{\sigma}{\epsilon}$ the state reached when applying $\epsilon$ in $\sigma$.

We assume that a bound on the number of objects for both training and testing problems is known. This is needed to construct the size of the first layer of the feed-forward neural network that we will learn. Hence, we define the training set as follows.
\begin{definition}
  The \deftitle{set of $k$-bounded planning problems} for a domain $\mathcal{D}$ is $\mathbb{P}_\mathcal{D}^k \eqbydef \{\tuple{\mathcal{O}, \mathcal{I}, \mathcal{G}} \mid k \ge |\mathcal{O}|\}$.
\end{definition}

\noindent
We are interested in a particular subset of k-bounded planning problems $\dataset \subseteq \mathbb{P}_\mathcal{D}^k$ and we assume a uniform distribution over it. 
In the following, we assume to have a sample of this distribution available as a subset $\mathcal{P}_\mathcal{D}^k \subseteq \dataset$ which will be our training set. 
Our aim is to synthesize planning guidance that can aid the search of a planner for problems drawn from the same distribution of $\mathcal{P}_\mathcal{D}^k$.

The most common form of planning guidance is a heuristic function. The heuristic takes in input a search state and the description of the problem being solved (i.e. it takes the state of the search, the goal formulation and the set of objects) and returns (an estimation of) the number of events needed to reach a goal state from the input state. 
In this work, we focus on satisficing planning and therefore do not consider optimality metrics.
\begin{definition}
  The \deftitle{optimal distance heuristic} for a training set $\mathcal{P}_\mathcal{D}^k$ is a function
  $
  \myhstar : \left( \bigcup_{\mathcal{P} \in \mathbb{P}_\mathcal{D}^k} \mathcal{S}_{\tuple{\mathcal{D}, \mathcal{P}}} \right) \times \mathbb{P}_\mathcal{D}^k \rightarrow \mathbb{R}
  $
  s.t. for each $\mathcal{P} \in \mathcal{P}_\mathcal{D}^k$ and each
  state $\sigma \in \mathcal{S}_{\tuple{\mathcal{D}, \mathcal{P}}}$, $d \eqbydef
  h^*(\tuple{\sigma, \mathcal{P}})$ is the minimum number such that there exists a sequence of events $\tuple{\epsilon_1, \ldots, \epsilon_d}$ and a sequence of states $\tuple{\sigma_0, \sigma_1, \ldots, \sigma_d}$ s.t. $\sigma_0 = \sigma$, $\sigma_d$ is a goal state, $\epsilon_j \in \appl{\sigma_{j-1}}$ and $\sigma_j=\successor{\sigma_{j-1}}{\epsilon_j}$ for all $j \in \{1, \ldots, d\}$.
\end{definition}   
\noindent

Our first aim is to automatically learn an approximation of
$\myhstar$, then in \cref{section:multiqueue} we will present another form of planning guidance we can learn, which is a ranking function. The overall idea is that the learned guidance will generalize to problems ``similar'' to the ones in $\mathcal{P}_\mathcal{D}^k$, i.e. problems in $\dataset \setminus \mathcal{P}_\mathcal{D}^k$.  


\section{MDP for Heuristic Synthesis }
\label{section:mdp}


In order to learn informed planning guidance, we encode our training problems as an MDP adapted from \cite{aaai21}, which we then use for reinforcement learning. In this MDP, the initial state is chosen uniformly at random from the set of initial states of the training planning problems (for generalization) and then the MDP state space is isomorphic to the search space of a temporal planner for the selected instance. 

Formally, given a bounded planning problem set $\mathcal{P}_\mathcal{D}^k$, its MDP encoding $\mymdp$ is the MDP 
defined as follows.
\begin{itemize}
\item The state space is $S \eqbydef \{\dummystate\} \cup  \bigcup_{\mathcal{P} \in \mathcal{P}_\mathcal{D}^k} \{ \tuple{\sigma, \mathcal{P}} |  \sigma \in \mathcal{S}_{\tuple{\mathcal{D}, \mathcal{P}}}  \}$,  and it is infinite since the planning state space is infinite (\dummystate is a fresh symbol denoting a special state whose purpose will become clear in the following);
\item the action space is $A \eqbydef \{\dummyaction\} \cup \bigcup_{\mathcal{P} \in \mathcal{P}_\mathcal{D}^k} \mathcal{E}_{\tuple{\mathcal{D}, \mathcal{P}}}$ (where \dummyaction is another fresh symbol for a special action);
\item the probability distribution over initial states is $$\initdistrib(\tuple{\sigma,\mathcal{P}}) \eqbydef \begin{cases}
\frac{1}{|\mathcal{P}_\mathcal{D}^k|} & \mbox{if } \sigma \mbox{ is the initial state for } \mathcal{P} \\ 
0 & \mbox{otherwise;} \\
\end{cases}$$
\item the transition function is deterministic and is $$T(\tuple{\sigma,\mathcal{P}}, a) \eqbydef
\begin{cases}
  \tuple{\sigma', \mathcal{P}} & \mbox{if } a\not=\dummyaction \mbox{ and } \sigma' = \successor{\sigma}{a} \\ 
  \dummystate & \mbox{if } a = \dummyaction \mbox{ and } \tuple{\sigma, \mathcal{P}} \in \deadends \\ 
\end{cases}$$ where $\deadends \eqbydef \{ s \in S | s = \tuple{\sigma, \mathcal{P}} \mbox{ and } \appl{\sigma} = \varnothing\}$ is the set of dead ends (states without applicable events) for the planning problems;

\item the terminal states of the MDP are $\goalstates \cup \{\dummystate\}$  where $\goalstates \eqbydef \{ s \in S | s = \tuple{\sigma, \mathcal{P}} \mbox{ and } \sigma \mbox{ is a goal state for } \mathcal{P}\}$ is the set of goal states of the planning problems;
\item the reward function $R(s,a,s')$ and the discount rate $\gamma$ will be defined in the next section, because there are multiple possibilities. 
\end{itemize}
\noindent

%
The overarching idea is to use RL to estimate the optimal value function $V^*_{\mymdp}$ and from that estimation, derive an estimation of $\myhstar$ (or other forms of guidance).
For this purpose, any RL algorithm would work, but we use the same RL algorithm as in the base framework, which is an adaptation of value iteration with replay buffer.  
We borrow from \cite{aaai21} the fixed-size vector representation of the MDP state to be taken as input to our neural network, in order to allow for a fair comparison.
The vectorization of an MDP state $\tuple{\sigma, \mathcal{P}}$ with $\mathcal{P}=\tuple{\mathcal{O}, \mathcal{I}, \mathcal{G}}$ contains both the encoding of the planner search state $\sigma$ and the encoding of the problem $\mathcal{P}$. In particular, the encoding of $\sigma$ contains not only the predicate values but also the temporal aspects including a digest of the STN $\sstn$, as per \cref{def:search-state}. The encoding of $\mathcal{P}$ consists in a vectorization of the set of objects $\mathcal{O}$ and a vectorization of the set of goals $\mathcal{G}$.

From now on, we will simply write $V^*$ in place of $V^*_{\mymdp}$, $h^*$ in place of $\myhstar$ and $\mathcal{M}$ in place of $\mymdp$. 


\subsection{Binary Reward Schemata} \label{section:bin-reward}

\paragraph{Theoretical Formulation.} \label{section:old-theo}
To formulate our MDP, a possible choice of reward function (called the ``binary reward'') is: 
\begin{equation*}
     \oldr(s,a,s') \eqbydef
    \begin{cases}
      1    & \mbox{if $s' \in \goalstates$} \\
      -1    & \mbox{if $s \in \deadends$} \\
      0    & \mbox{otherwise.}
    \end{cases}
\end{equation*}
The MDP using this reward is indicated as $\myoldmdp$ and is discounted with factor $0<\gamma<1$.
Moreover, its optimal value function $\oldv^{*}$ is in the range $[-1,+1]$: it is positive for ``good'' states (states leading to a goal) and negative for states from which goals are unreachable (i.e., dead-end states or states that lead to an infinite branch with no goal states).  
With this choice, \cite{aaai21} proved that the relationship between the optimal value function for the MDP $\oldv^{*}$ and the optimal heuristic function for the training planning problems is
\begin{equation}\label{eq:oldr_hstar}
h^*(s) = 
\begin{cases}
    \log_{\gamma}(\oldv^{*}(s))+1 & \mbox{if $\oldv^{*}(s) > 0$} \\
    +\infty                     & \mbox{otherwise}
\end{cases}
\forall s \in S \setminus \{\dummystate\}.
\end{equation}
Intuitively, a state $s$ whose distance from the nearest goal is $d$ has $h^*(s) = d$ and optimal expected discounted return $\oldv^{*}(s) = \gamma^{d-1}$. 

\paragraph{Practical Adjustments.} \label{section:old-prac}
When estimating $\oldv^{*}$ with a neural network (or any other function approximator) $V^{nn}_{bin}$, we make some practical adjustments.   
Since the MDP state space is potentially infinite, learning episodes are truncated after $\deltarl$ steps ($\deltarl$ is a hyperparameter) and the target for the value function update at episode truncation (we will call this ``truncation bootstrap'') is
%
\begin{equation} \label{eq:boostrap_oldr}
    V^{nn}_{bin}(s_{t+1}) \eqbydef 0 \quad \mbox{when } t=\deltarl 
\end{equation}
where $s_{t}$ is the state visited at time $t$. 
Here, value zero is chosen as a target because it is lower than the value of all ``good'' states and higher than the value of all states from which any goal is unreachable.


Due to episode truncation, the transformation function from the value function to the heuristic is adjusted in the following way.
\begin{equation} \label{eq:ugly_formula}
    \oldh^{nn}\!(s) \eqbydef\!\!
    \begin{cases}
      \min(\log_{\gamma}(\oldv^{nn}(s))+1, \deltahold)              & \mbox{if $V^{nn}_{bin}(s)\! >\! 0$} \\
      \deltahold                                                   & \mbox{if $V^{nn}_{bin}(s)\! = \!0$} \\
      2 \deltahold \!\!-\!\! \min(log_\gamma(-\oldv^{nn}\!(s)), \deltahold) & \mbox{otherwise}
    \end{cases}
\end{equation}
where $\deltahold \geq \deltarl$ is a planning parameter.
We cap the heuristic of states with positive $\oldv^{nn}$ to \deltahold because we want to always prefer a state from which the goal is reachable albeit far than a state whose value is uncertain due to episode truncation.
The capping also serves the purpose of attenuating numerical errors due to the logarithm.
Negative values are mapped to finite, although large, heuristic values because we cannot reliably say that a state with a learnt negative value is a dead end since learning can be imperfect.
\deltahold represents the depth to which we expect the learned value function to be informative. Since the search space is a tree, the RL algorithm can learn up to a distance \deltarl from the initial state. We can have $\deltahold \geq \deltarl$ if the neural network is able to generalize beyond that depth. The value function of ``good'' states is roughly in $[\gamma^{\deltahold}, 1]$ and the value function of ``bad'' states is in $[-1, -\gamma^{\deltahold}]$; values in $[-\gamma^{\deltahold}, \gamma^{\deltahold}]$ are ``uncertain'' i.e. they are not reliable estimates.
The hyperparameter $\deltahold$ needs to be carefully chosen: if it is too small, we lose information due to the capping; too large values may provide heuristic values that are too large compared to the $g$ (distance from initial state) values in the $A^*$ algorithm.

\subsection{Counting Reward Schemata} \label{section:cnt-reward}
\paragraph{Theoretical Formulation.} \label{section:new-theo}
An alternative choice of reward function, which we call the ``counting reward'', is the following.  
\begin{equation} \label{eq:counting_reward}
\newr(s,a,s') \eqbydef -1 \quad    \forall s,s' \in S, a \in A  
\end{equation}
and the MDP $\mynewmdp$ is undiscounted, i.e. $\gamma=1$. 
%
We also need to adapt the MDP transition relation to accommodate dead ends: we add a self-cycle on state $\dummystate$ (which gives reward $-1$ as all other transitions) and no longer consider it as a terminal state.
The optimal value function will be in the range $\newv^* \in (-\infty,0)$ for ``good'' states and $\newv^*=-\infty$ for ``bad'' states and the mapping from the value function to the heuristic becomes:
\begin{equation} \label{eq:beautiful_formula}
h^*(s) = - V^*_{cnt}(s) \quad    \forall s \in S \setminus \{\dummystate\}.
\end{equation}

We do not provide a proof of \cref{eq:beautiful_formula} for space reasons. It is well-known that in RL environments modeling unit cost graphs the reward function (\ref{eq:counting_reward}) provides shortest paths; ours is a generalization for infinite trees. 

The advantage of this new reward schema is that we directly learn the heuristic function, which is the function we ultimately want for planning (solving the Bellman equation for the optimal value function with the counting reward is equivalent to solving the Bellman equation for $h^*$).
%
In addition, we avoid the numerical instability of the logarithmic transformation in \cref{eq:ugly_formula}. 


\paragraph{Practical Adjustments.} \label{section:new-prac}
The formulation needs to be adapted when using a function approximator $\newv^{nn}$ for estimating $V^*_{cnt}$.
The MDP $\mynewmdp$ described in the previous paragraph is not proper, i.e. not all policies lead to a terminal state, since there are infinite branches of the search tree (as already observed in \cref{section:old-prac}) and there are the self-loops in dead ends. With $\gamma=1$ and an improper MDP, the optimal value function can diverge, and this is a problem for learning. To solve this issue, we truncate episodes after \deltarl steps (like we do for the binary reward) and  
we slightly change the MDP encoding for dead ends: we remove the self-cycle on $\dummystate$ and consider it again as a terminal state, and we provide a big negative reward on dead ends:
\begin{equation*}
     \newr(s,a,s') \eqbydef
    \begin{cases}
      -2 \deltahnew    & \mbox{if $s \in \deadends$} \\
      -1    & \mbox{otherwise;}
    \end{cases}
\end{equation*}
where $\deltahnew$ is a learning parameter.
When $\deltahnew \rightarrow \infty$ this modified MDP becomes equivalent to $\mynewmdp$.


The target for the value function at episode truncation is:
\begin{equation} \label{eq:bootstrap_newr}
    \newv^{nn}(s_{t+1}) \eqbydef - \deltahnew \quad \mbox{when } t=\deltarl .
\end{equation}

The range of possible values for
$\newv^{nn}$ is $[-\deltarl,0]$ for ``good states'', $[- \deltahnew - \deltarl, -\deltahnew]$ for ``uncertain'' states (due to episode truncation), $[-2\deltahnew-\deltarl,-2\deltahnew]$ for ``bad'' states.
Notice that if the parameter $\deltahnew$ is chosen so that $\deltahnew<\deltarl$, then the range of ``uncertain'' states partially overlaps with the other two ranges (in particular, it overlaps with the far ``good'' states). We need to choose $\deltahnew$ neither too big otherwise the uncertain states will hardly ever be extracted from the queue in $A^*$ (and same for the bad states which are not guaranteed to be bad since learning is imperfect), nor too small otherwise bad states will be mixed with good states. 

A heuristic is extracted from $\newv^{nn}$ as in the theoretical case:
\begin{equation} \label{eq:practical_beautiful_formula}
\newh^{nn}(s) \eqbydef - \newv^{nn}(s) \quad  \forall s \in S.
\end{equation}

We highlight that the counting reward schema is specific for temporal planning domains. The fact that in temporal planning the search space is a tree that we truncate allows us to set $\gamma=1$ and obtain the relationship (\ref{eq:practical_beautiful_formula}).
In classical planning, instead, it is not possible to set $\gamma=1$ because there can be cycles in the graph of the state space. 

In summary, we considered two reward schemata, resulting in different optimal value functions $V^*$ for the MDP, but both providing the optimal heuristic $h^*$ if transformed properly. For both schemata we showed the theoretical MDP for which the relationship between $V^*$ and $h^*$ can be established.
When using function approximation to estimate $V^*$, we introduced some practical adjustments to account for episode truncation (and also to have bounded value function in the case of the counting reward): for the binary reward, the MDP is the same but the transformation function changes; for the counting reward, we change the MDP encoding (the reward and the self-loop) but we keep the same transformation function. 

\section{Exploiting Symbolic Heuristics} \label{section:hsym} 
In this section, we explore three different ways of leveraging symbolic heuristic functions to improve the performance of our learning and planning framework.

Hereinafter, we assume a domain-independent heuristic \symh is given (e.g. $h_{\mathit{ff}}$ or $h_{add}$): we do not require particular formal properties other than correctness on states from which goals are unreachable: if $\symh(s)=+\infty$, then $h^*(s) = +\infty$. Moreover, we can also use different heuristics for the following three techniques, if they are applied in combination.

In \cite{aaai21}, a symbolic heuristic is used in the behavior policy, i.e. for choosing the non-greedy action during learning: instead of choosing a random action as in the standard epsilon-greedy policy, an action is sampled with probability inversely proportional to the heuristic value.  
Another way to exploit the prior knowledge given by symbolic heuristics is to treat any state $s$ where $\symh(s)=+\infty$ like a dead end: the action $\dummyaction$ is added to $s$ and leads to $\dummystate$ giving the dead end reward, and all applicable events in $s$ are removed from the MDP. 

\subsection{Truncation Bootstrap} \label{section:bootstrap}
Our first proposal to leverage symbolic heuristics consists in improving the truncation bootstrap i.e.  estimating the value function of states at the boundary of the subtree explored by the RL algorithm. Using $\symh$ yields a potentially better estimate than using a constant value like in Equations (\ref{eq:boostrap_oldr}) and (\ref{eq:bootstrap_newr}). Concretely, we use the following targets when $ t=\deltarl$.
\begin{equation*}
    \oldv^{nn}(s_{t+1}) \eqbydef
    \begin{cases}
        \gamma^{\symh(s_{t+1})-1} & \mbox{if } \symh(s_{t+1})<+\infty\\
        -1 & \mbox{otherwise} 
    \end{cases}
\end{equation*}
\begin{equation*}
    \newv^{nn}(s_{t+1}) \eqbydef
    \begin{cases}
        -\symh(s_{t+1}) & \mbox{if } \symh(s_{t+1})<+\infty\\
        -2 \deltahnew & \mbox{otherwise} 
    \end{cases}
\end{equation*}
%
%
Notice that this implies that the concept of ``uncertain'' interval of values disappears, in the sense that the value function of such states falls in either the ``good'' or the ``bad'' interval depending on the symbolic heuristic. 

We highlight that the truncation bootstrap is more relevant in temporal planning than in classical planning because the search tree is unexplored beyond depth $\deltarl$. In classical planning, one might use the value estimated by the neural network at $s_{t+1}$ as target for updating the value at $s_{t}$, while we empirically observed that this does not work for temporal planning.

\subsection{Residual Learning} \label{section:residual}

The aim of \cite{aaai21} is to learn the optimal heuristic function eagerly from scratch; instead, we propose to exploit $\symh$, which already tries to approximate $h^*$, as a basis on top of which we learn a residual. Concretely, we learn an additive residual $r^{nn}$ (dependent on the reward schema) of the optimal value function with respect to the symbolic heuristic. 
The intuition is that the function to learn $r^{nn}$ will be more regular than $V^{nn}$ since part of the complexity of $h^*$ is already modeled by $\symh$, and thus the learning task will be simplified.


With the $\myoldmdp$ formulation (\cref{section:old-theo}), we set
\begin{equation*} 
\oldv^{*}(s) = \oldres^{*}(s) + \oldphi(s)
\end{equation*}
where $\oldres^{*}$ represents the learnable part of the value function and $\oldphi$ is the fixed symbolic part defined as:
\begin{equation} \label{eq:phi}
    \oldphi(s) \eqbydef
    \begin{cases}
        \gamma^{\symh(s)-1} & \mbox{if } \symh(s)<+\infty\\
        -1  & \mbox{otherwise.} 
    \end{cases}
\end{equation}
Intuitively, \cref{eq:phi} is essentially the inverse of \cref{eq:oldr_hstar}.


As per the following proposition, if the $\symh$ were perfect, then the perfect residual would be the constant zero function, except for states with negative value. This is because $h^*$ does not estimate the distance to dead ends whereas $\oldv^*$ does.
\begin{proposition}  
    If  $\symh=h^* $ then $ \oldres^{*}(s) = 0  \quad \forall s \in \{s \in S |  \oldv^{*}(s)>0  \} $.
\end{proposition}  
\noindent
Since the symbolic heuristic is in reality an approximation of $h^*$, the residual will be either positive or negative depending on whether $\symh$ underestimates or overestimates $h^*$.
In practice, we set
$
\oldv^{nn}(s) \eqbydef \oldres^{nn}(s) + \oldphi(s)
$
where $\oldv^{nn}$ is the neural network output which estimates the correction of \oldphi.

Analogously, with the counting reward on $\mynewmdp$, the following equations and proposition hold. 
\begin{align*} 
    \newv^{*}(s) &= \newres^{*}(s) + \newphip(s)\\
    \newphip(s) &\eqbydef - \symh(s) \quad \forall s \in S
\end{align*}

\begin{proposition}
    If $ \symh=h^* $ then $ \newres^{*}(s) = 0 \quad \forall s \in  S$.
\end{proposition}

With $\newr$, learning a residual of the value function is equivalent to learning a residual of the heuristic function: thanks to \cref{eq:beautiful_formula}, $h^*(s) = \symh(s) - \newres^{*}(s)$.
With $\oldr$, instead, we learn a residual of the value function, not of the heuristic function directly and we need to account for the inverse of the logarithmic transformation. 

For the ``practical'' MDP using $\newr$, we set
\begin{align*} 
\newv^{nn}(s) &\eqbydef \newres^{nn}(s) + \newphi(s)\\
\newphi(s) &\eqbydef
\begin{cases}
    \newphip(s)  & \mbox{if } \symh(s)<+\infty\\
    -2 \deltahnew  & \mbox{otherwise.} 
\end{cases}
\end{align*}


Instead of viewing learning a residual as a way of fixing part of the value function estimation thanks to a symbolic heuristic, there is an alternative interpretation proposed by \cite{rl-classical}. The authors show that learning an additive residual of the value function with respect to some potential function (computed from the symbolic heuristic) is equivalent to learning the full value function of the MDP when adding a term to the reward function which depends on the potential function.  
Therefore, learning a residual can be seen as providing a dense reward computed from an existing heuristic.
This can mitigate the issue of sparse reward which is inherent of the MDP encoding of planning, and accelerate the reinforcement learning process since the agent reaches goals more quickly during training. 
The reward function adopted by \cite{rl-classical} is similar to our counting reward, with the difference that they set $\gamma<1$. For temporal planning we can set $\gamma=1$ since the search space is a tree that we truncate at depth $\deltarl$, whereas in classical planning there can be regions of the search space where the RL agent loops indefinitely without reaching a goal nor a dead end and thus the MDP is not proper. As already presented, setting $\gamma=1$ gives us the property that learning a residual of the value function is the same as learning a residual of the heuristic; instead, they need to use a transformation function on the heuristic function due to discounting, like we do for the binary reward.  

\subsection{Multiqueue and Ranking Function} \label{section:multiqueue}

Given the learned value function, \cite{aaai21} transforms it into a heuristic that is used in a weighted $A^*$ search schema. Here, we propose an alternative approach to exploit the learned value function for planning.


The overall idea is to adopt a multiple-queue planning technique \cite{multiqueue} to balance learned information with systematic search guided by a symbolic heuristic $\symh$. The learned information is in the form of a \emph{search state ranking} that is a function that can order any pair of search states, and it is used in a Greedy Best-First Search schema. 
Concretely, we directly use the value function synthesized by the RL algorithm as a means of ranking the search states generated by the planner, and we use a round-robin approach to either pick from the queue of states ordered by the standard planning heuristic or from the queue ordered according to the learned ranking.

\begin{definition}
  \label{def:learning-problem}
Given a bounded planning problem set $\mathcal{P}_\mathcal{D}^k$, learning a search-state ranking means approximating a function $\rank: \bigcup_{\mathcal{P} \in \mathbb{P}_\mathcal{D}^k} \mathcal{S}_{\tuple{\mathcal{D}, \mathcal{P}}} \times \mathbb{P}_\mathcal{D}^k \rightarrow \mathbb{R}$ such that for any pair of search states $s,s' \in \mathcal{S}_{\tuple{\mathcal{D}, \mathcal{P}}}$ for some $\mathcal{P} \in \mathcal{P}_\mathcal{D}^k$, if $\myhstar(s) < \myhstar(s')$, then $\rank(s) < \rank(s')$.
\end{definition}

A search-state ranking function $u$ for two search states $s$ and $s'$ such that $u(s) < u(s')$ indicates that the ranking regards $s$ as a more promising state than $s'$. Note that the absolute value of a ranking function $u(s)$ gives no information if not compared with another value of the ranking function, like $u(s')$. In this sense, a ranking function is less versatile than a heuristic and cannot be directly used in a search schema like $A^*$. However, we can use a learned search-state ranking function $u$ in a multi-queue search schema as shown in \cref{alg:search}. The algorithm hybridizes a (weighted) $A^*$ queue $Q[0]$, ordered according to \symh, and a $GBFS$ queue $Q[1]$, ordered according to $u$.
In the algorithm, the state expansion alternatively selects from either queue and removes the extracted element from both queues; moreover, each successor state is added to both queues.
This allows us to combine the greedy exploitation of the learned information with an explorative, systematic search.

\begin{algorithm}[tb]
\caption{Multiqueue search algorithm}
\label{alg:search}
\scriptsize
\algsetstretch
\begin{algorithmic}[1]
  \Procedure{Search}{$w$}
    \State{$I \gets $ \Call{GetInit}{\ }; \ \ $ g(I) \gets 0$; \ \ $i \gets 0$}
    \State{$Q \gets [$ \Call{NewPriorityQueue}{\ },  \Call{NewPriorityQueue}{\ } $]$}
    \State{\Call{Push}{$Q[0]$, $I$, $\symh(I)$}} \Comment{$Q[0]$ is sorted as a weighted $A^*$ open list}
    \State{\Call{Push}{$Q[1]$, $I$, $u(I)$}} \Comment{$Q[1]$ is sorted as a GBFS queue with ranking $u$}
    \While{$|Q[0]| > 0$}
      \State{$c \gets $\Call{PopMin}{$Q[i]$}}
      \State{$i \gets (i + 1) \mbox{ mod } 2$}
      \State{\Call{Remove}{$Q[0]$, $c$}; \ \ \Call{Remove}{$Q[1]$, $c$}} \Comment{Remove $c$ from both queues}
      \If{$c$ is a goal state}
  \Return{\Call{GetPlan}{$c$}}
\Else \textbf{ for all} $s \in$ \Call{Succ}{$c$} \textbf{do}
  \State $g(s) \gets g(c) + 1$
  \State \Call{Push}{$Q[0]$, $s$, $(1-w) \times g(s) + w \times \symh(s)$}
  \State \Call{Push}{$Q[1]$, $s$, $u(s)$}
\EndIf
    \EndWhile
  \EndProcedure
\end{algorithmic}
\end{algorithm}


For both our reward schemata, we can extract a search-state ranking function from the value function by simply inverting the value function itself. In symbols, $\rank_{bin}(s) = -\oldv^*(s)$ and $\rank_{cnt}(s) = -\newv^*(s)$. This fact is obvious for $\newr$, because $\rank_{cnt}(s) = -\newv^*(s) = h^*(s)$. For $\oldr$, it is easy to see from \cref{eq:oldr_hstar} that if $\oldv^*(s) < \oldv^*(s')$, then $h^*(s) > h^*(s')$ (because $\gamma < 1$).




Intuitively, the multi-queue approach can make planning more robust since the symbolic heuristic can compensate for some errors in the learnt value function: when learning has some local imperfections and drifts the search away from goals, the symbolic heuristic can help to escape from such regions of the search space.

\section{Implementation Details}

\paragraph{Neural Network Architecture.}
We use the same neural network architecture as \cite{aaai21} to allow for a fair comparison,
and the search states (as per \Cref{def:search-state}) are encoded into vectors of real numbers of fixed size. The only difference is the activation function in the output layer, due to the different range of the value function in the case of the counting reward and/or the residual.
Given the network output $x$ after the last linear layer, with the binary reward we apply the activation function $y(x)=\sigmoid(x)$, where $\sigmoid$ indicates the sigmoid function so that we obtain the output range $y \in [-1,+1]$. With the counting reward we apply the activation $y(x)=(\sigmoid(x)-1) \cdot \frac{3}{2} \deltahnew$ instead, so that $y \in [-3 \deltahnew, 0]$.
When learning a residual, in the case of the binary reward we change the activation to $y(x)=\frac{3}{2} \sigmoid(x) - \frac{1}{2}$ so that the output range becomes $y \in [-2,+1]$. In this way, when $\oldphi$ is close to 1 we can correct the value function towards negative values up to $-1$ with a residual $\oldres^{nn}$ of $-2$. Vice versa, when $\oldphi$ is $-1$, the value function can be corrected towards zero with a residual of up to $+1$. Notice that we need not have $\oldres^{nn}>+1$ because if $\oldphi$ is negative we are sure that $\oldv^{nn}$ is negative as well. 
In the case of the counting reward, a similar reasoning leads to consider an activation $y(x)=(2\sigmoid(x)-1) \cdot \deltahnew$ so that $y \in [-3 \deltahnew, + \deltahnew]$.


\paragraph{Learned Heuristic.}
When the neural network output represents a residual, the learned value function obtained by combining the residual with the symbolic part $\phi$ may not be in the correct range, and thus we truncate it as follows before transforming it into a heuristic for planning.
In the case of the binary reward, the value can be outside the range $[-1,+1]$ and so we clip values above $1$ or below $-1$ before applying \cref{eq:ugly_formula}.
In the case of the counting reward, we clip the positive values to $0$ before applying \cref{eq:practical_beautiful_formula}.
Another practical adjustment is that when $\symh(s)=+\infty$, we set $h^{nn}(s)=+\infty$ without computing the residual.
Finally, before computing the predicted value function of a successor state $s$ we always check whether $s$ is a goal state, and in that case we set $h^{nn}(s)=0$, since the neural network is not guaranteed to output zero value for goal states.



\section{Experiments}

For our experimental evaluation, we consider an adaptation of the two benchmark planning domains used in \cite{aaai21}: namely, the ``\majsp'' and ``Kitting'' domains, and of the ``MatchCellar'' IPC domain. All the domains considered are temporally expressive \cite{cushing_temporally_expressive}.
\majsp consists of a job-shop scheduling problem in which a fleet of moving agents transport items and products between operating machines.
In MatchCellar, a set of fuses have to be mended while light is provided by a match to be lit.
In Kitting, some robots have to collect several components distributed in different locations of a warehouse in order to compose a kit and then deliver it to a specific location synchronizing with a human operator.
We created 655 instances of \majsp, 725 of Kitting and 575 of MatchCellar.

We implemented the learning part of our framework in Python3 and the rollout simulator in Rust. We adopt the PyTorch framework \cite{pytorch} for training the neural networks. We use a self-implemented planner inspired from
\tamer \cite{tamer} and written in Rust.  
The learning process takes in input the training instances and outputs the trained value function as a neural network. In the learning algorithm we set the following hyperparameters: $\gamma = 0.99$ for $\oldr$, the maximum size of the replay buffer is 50K, the batch size is $1000$, the maximum depth for an episode is $\deltarl=200$ for \majsp, $100$ for Kitting and $150$ for MatchCellar (due to the different expected size of plans), the learning rate of the optimizer is $10^{-4}$, $\deltahnew=100$ for \majsp, $50$ for Kitting and $75$ for MatchCellar. In the planning algorithm, we set $w = 0.8$ for $A^*$ algorithm and \cref{alg:search}, $\deltahold=600$ for \majsp, $300$ for Kitting and $450$ for MatchCellar.

To measure the effectiveness of our framework, we performed a 5-fold cross validation: for each domain, we generated the set of ground instances and we randomly partitioned it into 5 equally sized subsamples. In turn, we use each subsample as testing data for the planning part, and the remaining 4 subsamples as training data for learning. We repeat the learning process twice, resulting in 10 total runs.

All the experiments have been conducted on a AMD CPU EPYC 7413 2.65GHz; we imposed a 600s/20GB time/memory limit for executing all the planning approaches.
The learning algorithm has been executed for 25K, 50K and 100K episodes and the total training time was approximately equivalent to 225 days of computation on a single-core machine.
Benchmarks, code and all the plots are available in the additional material of this paper~\cite{AdditionalMaterial}.

We experimented with eight learning techniques, corresponding to the combinations of using the binary reward (\cref{section:bin-reward}) or the counting reward (\cref{section:cnt-reward}), bootstrapping at episode truncation with a constant value or with the symbolic heuristic (\cref{section:bootstrap}), and learning a residual or learning from scratch (\cref{section:residual}). For each of these eight learning techniques, we experimented with the two planning techniques of $A^*$ with $h^{nn}$ or the multi-queue (\cref{section:multiqueue}). We used $h_{\textit{ff}}$ as symbolic heuristic for all the three techniques of \cref{section:hsym}. We also compare against the baseline performance of our planner equipped with $h_{\textit{ff}}$ without learned information.

\begin{figure}[tb]
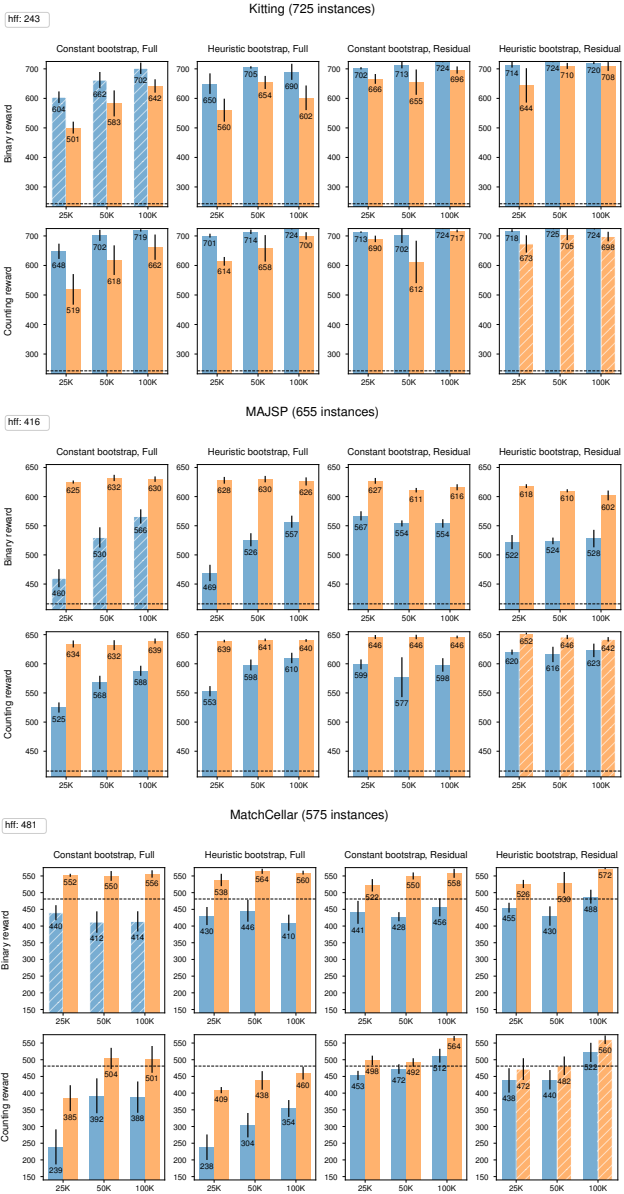

\centering
\resizebox{.965\columnwidth}{!}{\input{images/histogram_table_Kitting.pgf}}
\resizebox{.965\columnwidth}{!}{\input{images/histogram_table_MAJSP.pgf}}
\resizebox{.965\columnwidth}{!}{\input{images/histogram_table_MatchCellar.pgf}}
\caption{\label{fig:coverage}Experimental coverage}
\end{figure}
\begin{figure}[tb]
\centering
\includegraphics[width=.965\columnwidth]{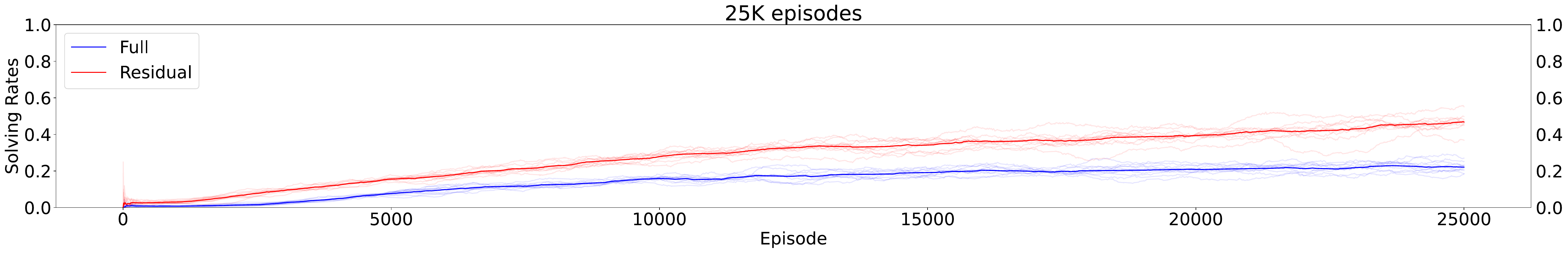}
\includegraphics[width=.965\columnwidth]{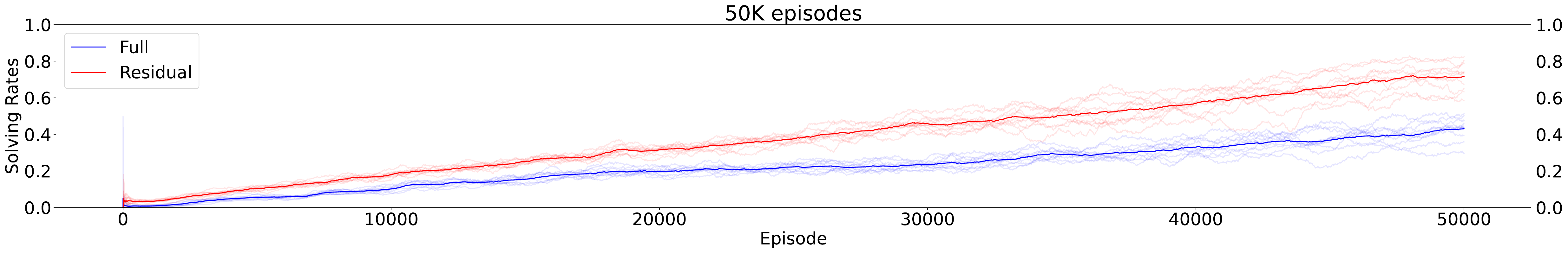}
\includegraphics[width=.965\columnwidth]{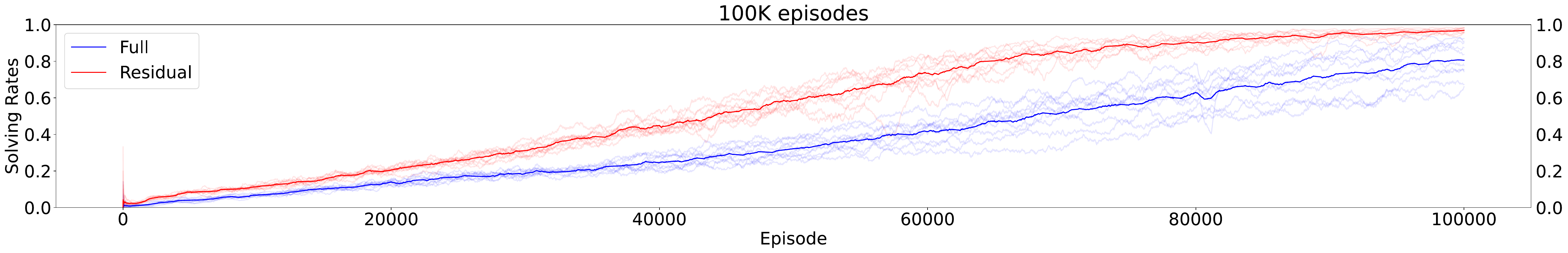}%
\caption{\label{fig:curves}Learning curves for Kitting, counting reward, heuristic bootstrap: residual (in red) vs. full (in blue). Y-axis is the fraction of episodes that reached a goal in the previous 1000 episodes. We plot a semi-transparent line for each cross-validation set and each run; the bold line is their average.}
\end{figure}

\cref{fig:coverage} reports the coverage results for our three domains.
The figure is read like a table with eight cells whose rows correspond to the configurations binary reward vs. counting reward, and whose columns are the combinations of the techniques truncation bootstrap with a constant value or with the heuristic, and learning from scratch (denoted with ``Full'') or learning a residual. In each of the eight cells, the x axis is the number of episodes, the y axis is the coverage. The vertical lines on top of the bars represent the standard deviation, computed by considering the two runs for each set. 
The blue bars are the coverage obtained with the single queue algorithm and orange bars are obtained with the multi queue algorithm.
The horizontal dashed line is the $h_{\mathit{ff}}$ coverage, whose value is also reported in the top left corner of the figure. In the Kitting and \majsp domains, all combinations of techniques are able to beat the fully symbolic planner.
The striped bars represent the coverage of the baseline which is a re-implementation of the approach by \cite{aaai21}  (top leftmost cell) and the coverage obtained when all our contributions are applied (bottom rightmost cell).  
Remarkably, the latter is consistently higher than the former for all number of episodes in the case of \majsp and MatchCellar, and comparable in the case of Kitting.

We now analyze the single techniques in isolation.
The counting reward performs better than the binary reward in Kitting and \majsp in almost all cases; in the MatchCellar domain the situation is more scattered.
The impact of the heuristic bootstrap is not decisive but positive in many cases.
The residual techniques perform better than the eager techniques across the spectrum.
It is important to note that when learning a residual, the planner is able to reach a good performance with a much lower number of episodes, hence reducing training time.
This can be also seen from the learning curves, which plot the fraction of problems solved during training: learning a residual is much faster than learning from scratch. 
Learning curves for Kitting with counting reward and heuristic bootstrap are reported in Figure \ref{fig:curves}. All other learning curves can be found in the additional material.


In the \majsp and MatchCellar domains, the multiqueue approach consistently performs better than the single queue for all combinations of techniques. In Kitting, instead, it performs worse; this could be due to the fact that the single queue coverage is already very close to the perfect coverage and the computation of the symbolic heuristic just adds overhead at planning time; furthermore, $h_{\mathit{ff}}$ alone is very weak in the Kitting domain, therefore the combination of learned heuristic with symbolic heuristic brings no benefit.

\section{Related Work}

Nowadays, combining automated planning with Machine Learning (ML) techniques is a very hot topic. 
However, with the exception of \cite{aaai21}, which we already discussed at length, and \cite{buffet2009factored}, all the other works we are aware of focus on classical planning.
ML can be used for different purposes in automated planning.
Some authors (e.g. \cite{model_learning,geffner_model_learning}) tackle the problem of
learning symbolic planning models from data; others (e.g. \cite{asnets,geffner-gnn,rossetti_icsps24}) aim to learn generalized policies. Furthermore, \cite{scala_plan_cost} aims at estimating the cost of a plan from instance features, while \cite{combine_heuristics} learns to compose multiple symbolic heuristics.
In \cite{buffet2009factored} a factored policy is learned with policy gradient RL for solving concurrent probabilistic temporal planning problems.

In this paper, we focus on learning heuristics for search-based temporal planners.
In this area, several papers perform supervised learning from a collection of plans to regress a heuristic function.
\cite{case-based-learning-heuristics} exploit a case-based database to construct planning heuristics, while learning of control policies for classical planning is studied in \cite{fern-ck}.
In \cite{learn_heuristics}, the search spaces generated by a classical planning heuristic are used to learn an incrementally better one. 
\cite{malte_learning} showed a comprehensive hyper-parameter experimentation for supervised learning of a classical planning heuristic. 
More recently, both Graph Neural Networks \cite{dillon_aaai24} and classical machine learning \cite{dillon_icaps24} have been used to learn heuristics for classical and numeric planning from a database of example plans. 
Differently from all these works, which are limited to classical planning, we tackle expressive temporal planning with intermediate conditions and effects and provide a fully-automated technique to learn heuristics from simulations via RL.

Similarly to our work, but differently in terms of expressiveness, \cite{rl-classical} propose the use of symbolic heuristics as dense reward generators that improve the sample efficiency of RL for classical planning. Their MDP formulation is similar to our counting reward schema with two key differences: first, it is discounted (because the search space in classical planning is not a tree as in temporal planning); second, they truncate the episode upon reaching a dead end, whereas we appropriately modify the MDP encoding to deal with dead ends. 

Finally, \cite{rank-learn} learn a classical planning state ranking function for use in a multi-queue planner. Differently from our work, they operate in a supervised setting, learning from a database of plans; instead we use RL to make the learning process fully automatic.

\section{Conclusion}

In this paper, we discussed how to exploit the information provided by a symbolic heuristic in a RL schema aimed at synthesizing guidance for search-based temporal planners. We use symbolic heuristics during learning, to mitigate the issues caused by episode truncation, and to learn a residual of an existing heuristic instead of learning the whole heuristic function from scratch; 
moreover, we exploit symbolic heuristics in planning to balance the greedy exploitation of the learned guidance with systematic search, in a multi-queue approach.

For future work, we plan to relax the assumption on the maximum number of objects by adapting recent results using Graph Neural Networks (GNN)~\cite{geffner-gnn} to the temporal planning case, and to experiment with multiplicative residuals instead of additive residuals.



\begin{ack}
This work has been supported by the STEP-RL project funded by the European Research Council under GA n. 101115870.
This work has been carried out while Irene Brugnara was enrolled in the Italian National Doctorate on Artificial Intelligence run by Sapienza University of Rome in collaboration with Fondazione Bruno Kessler.
\end{ack}



\bibliography{refs}

\end{document}